\documentclass[runningheads]{llncs}
\usepackage{graphicx}
\usepackage{amsfonts, amsmath, amsxtra, amssymb, latexsym,url}
\usepackage{makeidx}
\urldef{\mailsa}\path{berikov@math.nsc.ru}
\newcommand{\keywords}[1]{\par\addvspace\baselineskip
\noindent\keywordname\enspace\ignorespaces#1}
\usepackage[cp1251]{inputenc}
\usepackage[english]{babel}
\usepackage{multirow}

\usepackage{color}

\begin{document}

\title{Heterogeneous Transfer Learning in Ensemble Clustering \thanks{Submitted to the proceedings of the Second International Conference ''Situation, Language, Speech. Models and Applications'' (SLS 2019)}}

\titlerunning{Heterogeneous Transfer Learning in Ensemble Clustering}
\author{Vladimir Berikov\inst{1}\orcidID{0000-0002-5207-9764} }

\authorrunning{V. Berikov}

\institute{Sobolev Institute of mathematics,
Novosibirsk, Russia\\ 
\email{berikov@math.nsc.ru}\\
\url{http://www.math.nsc.ru}
}

\maketitle

\begin{abstract}
This work proposes an ensemble clustering method using
transfer learning approach. We consider a clustering problem, in
which in addition to data under consideration, ''similar'' labeled data are available. The datasets can be described with different features. The method is based on constructing
meta-features which describe structural characteristics of data, and their transfer from source to target domain. An experimental study of the method using Monte Carlo modeling has confirmed its efficiency. In comparison with other similar methods, the proposed one is able to work under arbitrary feature descriptions of source and target domains; it has smaller complexity.
\end{abstract}

\keywords{
clustering; transfer learning; ensemble of algorithms; co-association matrix}

\section{Introduction}
\noindent 
In machine learning, there is a fairly large number of models, methods and algorithms based on different approaches. Topical areas of research are {\it transfer learning} aimed at improving the quality of decisions by the usage of additional information from similar field of study, and {\it ensemble clustering} aspiring to increase the quality and stability of clustering results.

In transfer learning (closely related to {\it domain adaptation} and {\it knowledge transfer}), the basic idea is to use additional data (called {\it source data}) which is similar, in a certain sense, to the data of interest ({\it target data}). For example, one may use digital images of the same landscape but at other moments of time, or utilize data and results of text documents classification in a certain language, to documents written in another language.

Within this direction, there are various ways of setting the problem, for example, when one has (or has not) information about class labels in target and source domains; for the same
features (homogeneous domain adaptation) or different feature descriptions
(heterogeneous domain adaptation).
As a rule, it is assumed that the probability
distributions on target and source domains do not coincide. Transfer
learning methods are elaborated for pattern recognition problems, regression and cluster analysis \cite{Pan}. Different approaches to knowledge transfer have been developed, for example, based on the usage of source examples with some adjustable weights for constructing decision functions when predicting the target sample; searching for common feature descriptions; making use of the assumptions about the
coincidence of the distribution of some hyperparameters.

Ensemble clustering aims at finding consensus decision from multiple
partition variants \cite{Ghosh,Berikov14,BerPest,Boongoen}. As a
rule, this methodology gives robust and effective solutions,
especially in the case of uncertainty in the data structure. Properly
organized ensemble (even composed of ''weak'' algorithms)
significantly improves the overall clustering quality.

We consider a clustering problem, in which, in addition to the target dataset, one may use already classified dataset described by other features and belonging to another statistical population. A practical example is the segmentation of color image using a similar already segmented gray-scale image.

The problem of transfer learning with use of cluster ensembles was considered in \cite{Acharya}. In this work, the authors assume that source and target domains share common feature space and class
labels. The authors of \cite{Shi} suggest a general framework with
arbitrary feature descriptions. However, the proposed algorithm has a cubic time complexity with respect to the maximum sample size of source and target domains.

In this paper, we propose an ensemble clustering method using transfer learning approach. The idea of the method is as follows. On the first stage of analysis, both target and source data are examined independently using a cluster ensemble, in order to identify stable structural patterns. This results in obtaining meta-features describing structural data characteristics. On the second stage, the
relationships between the elements of coincide matrix of source data
and their meta-features are revealed with use of supervised
classification. On the next step, a transfer of the found
dependencies from source to target domain is performed and the
prediction of the coincidence matrix for target data is made.
Finally, on the basis of the obtained predictions, the final clustering partition of target data is obtained.

 In comparison with other similar methods, the proposed one is able to work under arbitrary feature descriptions of source and target domains;  it has smaller complexity.

In the rest of the paper, we give a detailed description of the proposed method and describe the results of its experimental investigation.

\section{Cluster analysis and knowledge transfer}
\subsection{Basic notation and problem statement}
\noindent Consider a set $ T=\{ a_{1},\dots, a_{N_{T}} \} $ of
objects from some statistical population. Each object is described by
a set of real-valued features $ X_{1}, \dots, X_{d_{T}} $. Through $
x=x (a)=(x_{1},\dots, x_{d_{T}}) $ we denote a feature vector for an
object $ a $, where $ x_{j}=X_{j} (a) $, $ j=1,\dots, d_{T} $, and
through $ X_{T} $ we denote data matrix $ X_{T}=(x_{i \, j}), \, \,
j=1, \dots, d_{T}, \, \, \, i=1, \dots, N_{T} $. It is required to
obtain a partition $ P=\{C_{1}, \dots, C_{K_{T}} \} $ of $ X_{T} $
on some number of $ K_{T} $ groups (clusters) in according to a
given quality criterion. The number of clusters can be preset
in advance or not; in this paper we assume the required
number is a fixed parameter.

Suppose there is an additional dataset $S=\{b_{1},\dots, b_{N_{S}}
\} $  where each object $b$ is described by real-valued features
$X'_{1} $, \dots, $ X'_{d_{S}} $,  and data matrix $ X_{S} $ is
given. A categorical attribute $ Y $ is specified, denoting a class to
which an object $ b \in S $ belongs: $ Y(b) \in \{1, \dots, K_{S} \}
$, where $ K_{S} $ is total number of classes. By classification
vector we understand a vector $ Y=\{y_{1},\dots, y_{N_{S}} \} $,
where $ y_{i}=Y (b_{i}) $, $ i=1,\dots, N_{S} $.

The set $ T $ is called target data, and the set $ S $ source data.
It is  assumed that $ X_{T} $ and $ X_{S} $ share some common
regularities in their structure, which can be detected by cluster
analysis and used as additional information when setting up the
desired partition of $ X_{T} $.

\subsection{Cluster ensemble}
\noindent Suppose we are able to create variants of the partitioning of $ X_{T} $ into clusters using some clustering algorithm $\mu $. The algorithm works under different parameter settings, or,
more generally, ''learning conditions'' such as initial centroids
locations, subsets of selected features, number of clusters or
random subsamples. On the $ l $th run it gives a partition of $
X_{T}$ on $ K_{l} $ clusters, $l=1,\dots, L $, where $ L $ is total number of runs.

For each pair of points $ a_{i} $, $ a_{j} \in X_{T} $ we define the
value $ h_{l}^{T}(i,j)=I \, [\mu _{l} (x_{i})=\mu_{l} (x_{j})] $,
where $ \mathbb{I} [\cdot] $ is an indicator function: $ \mathbb{I}
[true]=1 $; $ \mathbb{I} [false]=0 $, $ \mu _{l} (x) $  is an index
of the cluster assigned to a point $ x \in \, X_{T} $ by the
algorithm $ \mu $ on $l$ -th run. Let us calculate the averaged
co-association matrix $ H^{T}=(\bar {h}_{}^{T}(i,j)) $ with elements
$ \bar{h}^ {T}(i,j)=\frac{1}{L} \sum _{l=1} ^ {L} \,h_{l}^{T}(i,j)
$, $i,j=1, \ldots, N_{T} $.

The next stage is aimed at constructing the final partition of
$X_{T} $.  Elements of the matrix $ H^{T} $ are considered as measures
of similarity between pairs of objects. To form the partition, any
algorithm which uses these measures as input information can be
used. In this paper, we apply ensemble spectral clustering algorithm
\cite{BerNov} based on a low-rank decomposition of the averaged
co-association matrix which has near-linear time and storage
complexity.

The basic steps of the used ensemble clustering algorithm EC are
described below.

\vskip12pt \noindent\textbf{Algorithm EC:}

\noindent \textbf{Input}:

\noindent $ X_{T} $: target data;

\noindent $ L $: the number of runs of the ensemble clustering algorithm;

\noindent $ \Omega $: the set of algorithm's parameters;

\noindent $ K_{T} $: The required number of clusters in the
partition of $ X_{T} $ .

\noindent \textbf{Output}:

\noindent $ P=\{C_{1}, ..., C_{K_{T}} \} $: partition of $ X_{T} $.

\noindent\textbf {Steps}

\noindent 1. Get $ L $ variants of clustering partition for objects
from $X_{T} $ by randomly choosing the algorithm's parameters from
$\Omega $;

\noindent 2. Calculate the averaged co-association matrix $ H^{T} $
in low-rank representation;

\noindent 2. Using spectral clustering with low-rank represented
matrix $ H^{T} $ as input, find a final partitioning $ P=\{C_{1},
\dots, {K_{T}} \} $.

\noindent \textbf{end.} \vskip12pt

\subsection{Probabilistic properties of cluster ensemble}
\noindent Suppose we have \textit{iid} sample $X=\{x_{1}, \dots,
x_{N} \}$ generated from a mixture of $K$ distributions (classes).
Suppose there also exists a ground truth (latent, directly
unobserved) variable $Y$ that determines to which class an element
$x_i$ belongs: $Y_i \in \{1,\dots ,K \}$. Let $ Z(i,j)=I \, \,
[Y_{i} \ne Y_{j}] $, where $ i,j=1, \dots, N $. Let algorithm $\mu$
be randomized, i.e. it depends on a random set of parameters $\Omega
\in \mathbf{\Omega}$, and the sample partitions are formed using
independently selected statistical copies of $ \Omega $.

Each ensemble algorithm contributes to the overall collective
decision. Denote by $ v_{1}(i,j) $ the number of votes for the union
of $ x_{i} $, $ x_{j} $ into same cluster; and by $ v_{0}(i,j) $ the
number of votes for their separation. The value $ c(i,j)=I \, [\,
v_{1}(i,j)> v_{0}(i,j)] $ shall be called the ensemble solution for
$ x_{i} $ and $ x_{j} $ obtained in accordance with the voting
procedure. Conditional probability of classification error for each
pair is defined as $ P_{err}(i,j)=P [c(i,j) \ne Z(i,j) | X] $. The
following property was proved in \cite{Berikov17}.

\textbf {Theorem.} \textit {Let us suppose that for any $ i,j \, \,
\, \, (i \ne j) $ the symmetry condition is satisfied: $ P[h(i,j)=1
| Z(i,j)=1]=P [h(i,j)=0 | Z(i,j)=0]=q(i,j) $ where $ q(i,j) $ is the
conditional probability of correct decision. If the
condition of weak learnability $ 0.5 < q(i,j) \le 1 $ holds,
$ P_{err}(i,j) \to 0$ as $L$  approaches infinity.}

Therefore, under certain regularity conditions, the quality of  ensemble decisions improves with an increase in ensemble size.
However, in case of the violation of the assumptions, as well as
with a small number of ensemble elements, the quality of the
decisions can turn into a degenerate. To improve the ensemble
quality, it is possible to use information contained in the
additional (source) data.

\subsection{Usage of source data}
\noindent For source data $X_{S}$, classification vector $Y$ is
known. Therefore it is possible to calculate the coincidence matrix
$Z^{S}=\large( z^{S}(i,j)\large)$, where $z^{S}(i,j)= \mathbb{I}\,
\, [y_{i} = y_{j}]$, $i,j=1,\dots,N_{S}$. Despite the fact $X_{T}$
and $X_{S}$ belong to different statistical populations, we may
assume that some general structural regularities characterizing both
populations exist. The regularities can be found using cluster analysis. To get more robust results, we apply cluster ensemble algorithm independently for source and target data (for simplicity, we assume equal number $L$ of runs in the ensemble for each dataset). As the analysis proceeds, the averaged co-association matrix $ H^{S} $ is determined for $ X_{S}$.

For specifying the types of regularities under interest, one may use different approaches. In this paper, we consider characteristics of mutual positions of data points with respect to the found clusters and use them as meta-features describing common properties of source and target domains.

As the first type of meta-feature, we use frequencies of the assignment of object pairs to the same clusters (elements of matrix $H^S$). These values belong to interval $[0,1]$ and do not explicitly depend on initial feature dimensions.

Another meta-feature suggested in this work is based on Silhouette index which is defined for each data point and reflects its similarity to other points in the same cluster and dissimilarity to points from different clusters. Let $Sil(x_i)$, $Sil(x_j)$ denote Silhouette indices, averaged over ensemble partitions, respectively for points  $x_i,x_j \in X_{S}$. Denote $P ^{S}(i,j)=\frac{1}{2}(Sil(x_i)+ Sil(x_j))$; matrix $P ^{S}$ is determined for all pairs of points of source data. Similarly, matrix $P ^{S}$ is defined for target data.

Consider a problem of finding a decision function
\begin{equation}\label {Q0}
f: \; (P^{S}(i,j), H^{S}(i,j))  \mapsto Z^{S} (i,j)
\end{equation}
for predicting elements of $Z^{S}$ viewed as new class labels (0
or 1). A classifier can be found by usage of existing machine
learning algorithms such as logistic regression or support vector
machine. Then the found classifier can be transferred to target domain for
predicting $\hat{Z}^{T}=f(P^{T},H^{T})$.

The resulting coincidence matrix $\hat{Z}^{T}$ cannot be directly
used for clustering (for example, by finding connected components),
since it can lead to metric properties violation in feature space. For
example, if for some $i,j, k$, it holds $\hat{Z}^{T}(i,j)= 1 $ and
$\hat{Z}^{T} (i,k)= 1$, there is no guarantee that $\hat{Z}^{T}(j,
k) \ne 0$. For this reason, we propose to search for a partition by
solving the following optimization problem:
\begin{equation}\label {Q1}
\text{find } \hat{Z}^{*} \in \Psi_{K_T} :\, \, \hat{Z}^{*} = \arg
\min \, (\hat{Z}^{*} -\hat{Z}^{T})_{2}
\end{equation}
where $\Psi_{K_T}$ is a set of Boolean coincidence matrices
corresponding to all possible partitions of $ X_ {T} $ into $K_T$
clusters, $(\cdot)_{2}$ is the Frobenius norm of a matrix.

For an approximate solution, one may apply a procedure which starts
from some initial partition of $ X_{T}$ (in our implementation, found with  \textbf{EC} algorithm), and sequentially corrects
it by finding such points, which give the best improvement of functional in (\ref{Q1}) when migrating to another cluster. The iterations continue until the optimized functional becomes less than a given parameter $Q_{\min}$, or the
number of iterations (migrated points) exceeds the preset value
$\rm{It}_{\max}$.

The main steps of the proposed algorithm \textbf {TrEC} (Transfer Ensemble Clustering) are described below.

\vskip12pt \noindent\textbf{Algorithm TrEC:}

\noindent\textbf {Input}:

\noindent $ X_{T} $: target data;

\noindent $ X_{S} $: source data;

\noindent $ Y $: class labels for source data;

\noindent $ L $: number of runs for clustering algorithm $\mu$;

\noindent $\mathbf{\Omega_{S}}, \, \mathbf{\Omega_{T}}$: parameters
of algorithm $\mu$ working on $X_{S}$ and $X_{T}$, respectively;

\noindent $K_{T}$: required number of clusters in $X_{T}$.

\noindent \textbf{Output:}

\noindent $P =\{C_{1},\dots,C_{K_{T}} \}$: clustering partition of $X_{T}$.

\noindent \textbf{Steps:}

1. Generate $L$ variants of clustering for $X_{S}$ and $L$ variants of clustering for $X_{T}$ by random choice of parameters from $\mathbf{\Omega_{S}}, \,
\mathbf{\Omega_{T}}$, respectively;

2. Calculate co-association matrices $H^{T} , \,
H^{S}$ and matrices $P^{T}$, $P^{S}$, $Z^{S} $;

3. Find a decision function $f(P^{S},H^{S})$ for predicting elements
of $ Z^{S}$;

4. Calculate matrix $\hat{Z}^{T}=f(P^{T},H^{T})$;

5. Find matrix $\hat{Z}^{*}$ and corresponding partition $P$,
using the above-mentioned approximate procedure of solving (2);

\noindent\textbf{end.} \vskip12pt

In the implementation of \textbf{TrEC}, we use \textit{k}-means to design the
cluster ensemble. To find a solution to the problem (\ref{Q0}), we
apply Support Vector Machine (SVM). The overall complexity of the
algorithm with respect to sample size is $O(\max(N_{T},N_{S})^{2})$.

\section{Numerical experiment}
\noindent To verify the applicability of the suggested approach, we
have designed Monte Carlo experiments with artificial datasets. To
generate target data, we use the following distribution model. In
$24$-dimensional feature space, the two of modeled classes have
spherical form, and another two have strip-like form. First and
second classes are of Gauss distributions with unit covariance
matrix $N(\nu_i,\mathbf{I})$ , where $\nu_1=(0,\dots,0)^T$,
$\nu_2=(8,\dots,8)^T$. The coordinates of objects from other two
classes are determined recursively: $x_{k_{i+1}}=x_{k_i}+\gamma_1
\cdot \mathbf{1} + \varepsilon$, where $\mathbf{1}=(1,\dots,1)^T$,
$\gamma_1=0.2$, $\varepsilon$ is a Gauss random vector
$N(0,\gamma_2\cdot\mathbf{I})$, $\gamma_2=0.25$, $k=3,4$. For class
3, $x_{3_1}=(-6,6,\dots,-6,6)^T+\varepsilon$; for class 4,
$x_{4_1}=(6,-6,\dots,6,-6)^T+\varepsilon$. The number of objects for
each class equals 20.

Source data set has a similar structure, with the following
differences: feature space dimensionality equals $16$; the first and
second classes follow Gauss distribution $N(\lambda_{i},I)$,
$i=1,2$, where $\lambda_1=(0,\dots,0)^T$,
$\lambda_{2}=(6,\dots,6)^T$; the third and fourth classes are
determined as follows: $x_{k_{i+1}}^{'}=x_{k_{i}}^{'}+\gamma_{1}^{'}
\cdot \mathbf{1}+ \varepsilon'$, where $\gamma_{1}^{'}=0.2$,
$\varepsilon'$ is normally distributed vector $N(0,\gamma_{2}^{'}
\cdot \mathbf{I}) $, $\gamma_{2}^{'}=0.2$, $k=3,4$. For the third
class $x_{3_{1}}^{'} =(- 5,5,\dots,-5,5)^{T} +\varepsilon'$; for the
fourth class $x_{4_{1}}^{'}=(5,-5,\dots,5,-5)^{T}+\varepsilon'$. The
sample size for each class equals 25.

Examples of generated data are shown in figure 1.

\begin{figure}[ht!]
\centering
\includegraphics[width=\linewidth]{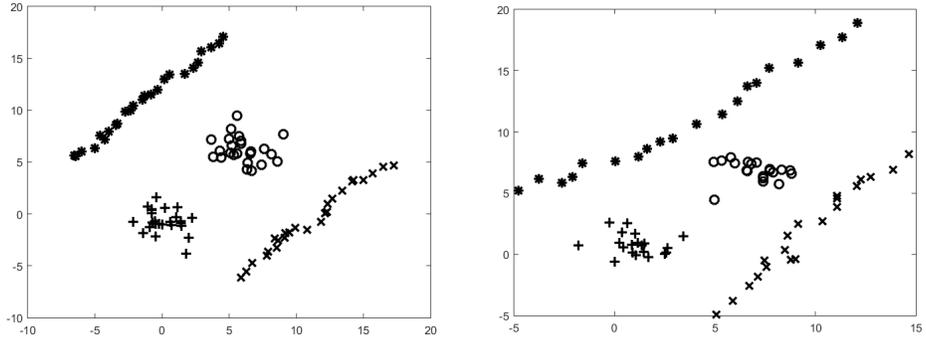} \caption{Examples
of sampled data (left: target data, right: source data); projection
on first two coordinate axes.} \label{fig1}
\end{figure}

To design the ensemble, we use random subspace method: each
clustering variant is built on three randomly selected features. The
number of elements in the ensemble equals 10; the number of clusters
$K_T=4$.

For training on $(P^{S}, H^{S})$, $\hat{Z}^{S}$ and predicting
elements of $\hat{Z}^{T}$ using $(P^{T}, H^{T})$, we apply SVM with the following
parameters: RBF kernel ($\sigma=4$), penalty parameter $C=10$.
Parameter $Q_{\min}=5$, $It_{\max}=40$.

In the process of Monte Carlo modeling, artificial datasets are
repeatedly generated according to the specified distribution model.

Figure 2 shows an example of SVM decision on the coordinate
plane with axes determined by values of $P^{S},P^{T}$ (horizontal
axis) and $ H^{S},H^{T}$ (vertical axis).

\begin{figure} [ht!]
\centering
\includegraphics[width=0.8\linewidth]{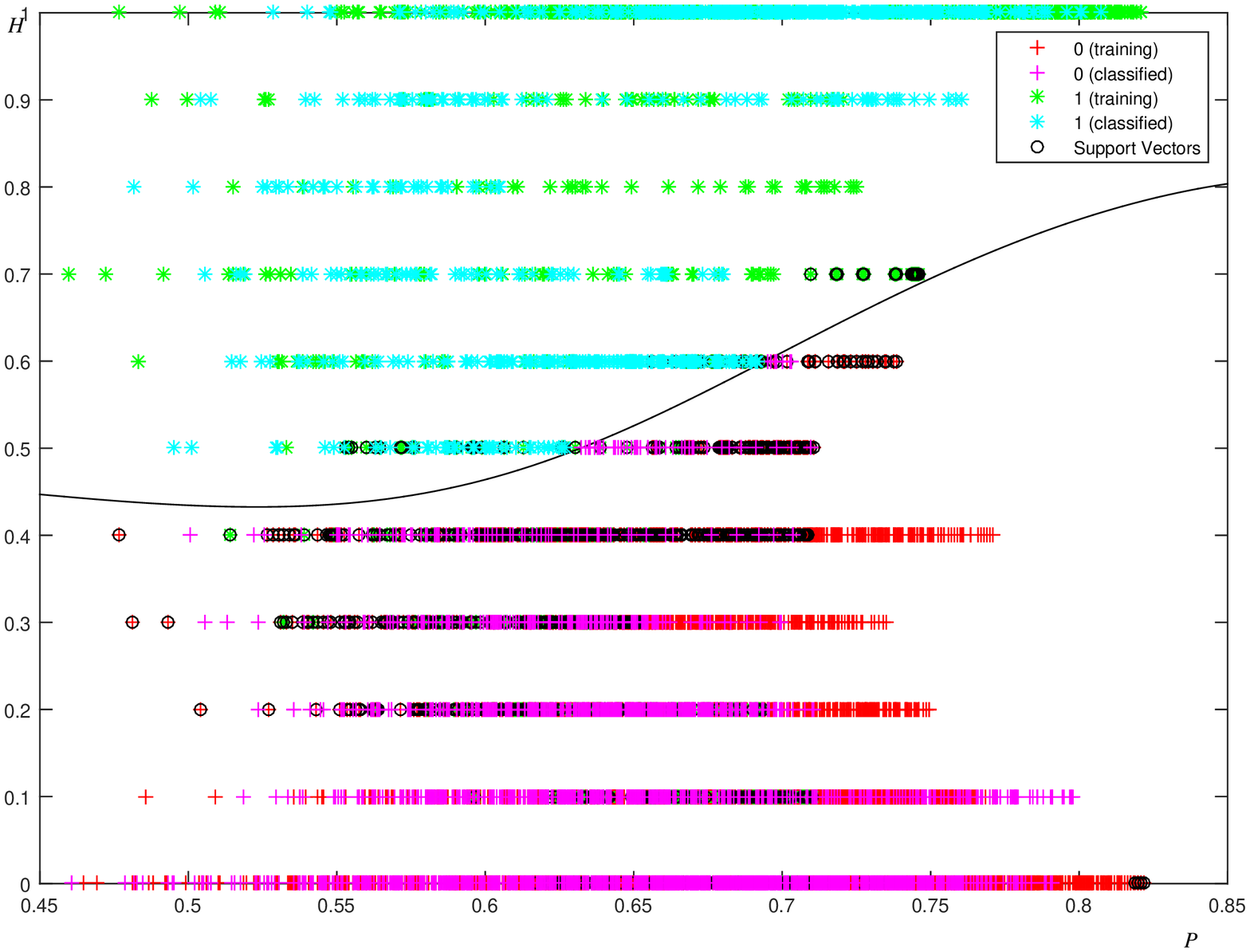} \caption{An example
of decision boundary.} \label{fig2}
\end{figure}

To evaluate the quality of clustering, we use
Adjusted Rand index $ARI$ \cite{Hub}.  The index varies from values around zero to $1$;
$ARI$ close to $1$ means a high degree of matching between the found
partition and the true one; $ARI$ close to zero indicates nearly
random correspondence. To increase the reliability of the results,
the accuracy estimates are averaged over 40 experiments. Two data processing strategies are compared: a) with use of the suggested
\textbf{TrEC} algorithm, and b) using \textbf{EC} algorithm, in which no transfer learning is utilized. The statistical analysis of the significance of the differences between the estimates is carried out using a paired Student's t-test.

As a result of the experiment, the following averaged quality
estimates are obtained: for \textbf{TrEC}: ${ARI} = 0.73$, and for
\textbf{EC}: $ARI = 0.56$. The paired Student's t-test shows
significant differences between the two estimates (p-value 0.0003).
Thus, despite the fact that the data distribution is quite difficult
for \textbf{k}-means (which is oriented on spherical-shaped clusters), the suggested method shows a statistically significant increase in decision quality.

We also estimate a significance of meta-features with respect to clustering quality. To this end, we try to exclude one of the features from the analysis and repeat the experiment. When co-association matrix-based meta-feature is used alone in \textbf{TrEC}, the averaged ${ARI} $ equals $0.68$. If only Silhouette-based meta-feature is employed, the averaged ${ARI} $ degrades to $0.31$. One can conclude from the experiment that the former meta-feature is more important than the latter, however, both of them  are useful in combination.

\section{Concluding Remarks}
This work has introduced an ensemble clustering method using
transfer learning methodology. The method is based on finding
meta-features describing structural data characteristics and their
transfer from source to target domain. The proposed method allows one to consider different feature sets describing source and target domains, as well as a different number of classes for both of them. The complexity of the method is of quadratic order, and is smaller than  the complexity of analogous algorithm.

An experimental study of the method using Monte Carlo modeling has confirmed its efficiency.

In the future, we plan to continue studying the theoretical properties of the proposed method and its further development aimed at faster
processing speed. Determining of useful types of meta-features, in addition to such as those proposed in the present work, is another important problem. A detailed comparison with existing combined ensemble clustering and transfer learning methods is our next objective. Application of the method in various
fields is also planned.

\section*{Acknowledgements}
The research was supported by RFBR grant 19-29-01175.


\begin{thebibliography}{99}

\bibitem{Pan}
Pan~S.,  Yang~Q. A Survey on Transfer
Learning. IEEE Transactions on Knowledge and Data Engineering. 22(10). 1345--1359 (2010)

\bibitem{Ghosh}
Ghosh J., Acharya~A. Cluster ensembles.
Wiley Interdisciplinary Reviews: Data Mining and
Knowledge Discovery. 1(5). 305--315 (2011)

\bibitem{Berikov14}
Berikov~V.~B. Weighted ensemble of algorithms for
complex data clustering. Pattern Recognition
Letters. 38. 99--106 (2014)

\bibitem{BerPest}
Berikov~V., Pestunov~I. Ensemble clustering
based on weighted co-association matrices: Error bound and
convergence properties. Pattern Recognition.
63. 427--436. (2011)

\bibitem{Boongoen}
Boongoen~T.,  Iam-On ~N. Cluster ensembles: A
survey of approaches with recent extensions and applications.
Computer Science Review. 28. 1--25. (2018)

\bibitem{Acharya}
Acharya~A. et al. Transfer learning with cluster ensembles. Proceedings of the 2011 International Conference on Unsupervised and Transfer Learning
workshop. Vol. 27. JMLR. org. 123--133. (2011)

\bibitem{Shi}
Shi ~Y. et al. Transfer Clustering Ensemble
Selection. IEEE Transactions on Cybernetics (Early
Access). 1--14. (2018)

\bibitem{BerNov}
Berikov~V.~B., Novikov~ I.~A. Model and
computationally efficient method of ensemble clustering.
Intelligent Data Processing: Theory and Applications: Book
of abstracts of the 12th International Conference. 10--11. (2018)

\bibitem{Berikov17}
Berikov ~V.~B. Construction of an optimal
collective decision in cluster analysis on the basis of an averaged
co-association matrix and cluster validity indices.
Pattern Recognition and Image Analysis. 27(2). 153--165. (2017)

\bibitem{Hub}
Hubert~L., Arabie ~P. Comparing partitions. Journal of Classification. 2(1). 193--218. (1985)

\end{thebibliography}
\end{document}